\documentclass{elektr}
\usepackage{hyperref}
\hypersetup{
colorlinks=true,
urlcolor=black,
citecolor=blue}
\usepackage[all]{xy,xypic}
\usepackage{amsfonts,amssymb,amsmath,amsgen,amsopn,amsbsy,theorem,graphicx,epsfig}
\usepackage{eufrak,amscd,bezier,latexsym,mathrsfs,eurosym,enumerate}
\usepackage[utf8]{inputenc}
\usepackage[english]{babel}
\usepackage{cleveref,multirow}
\usepackage[dvipsnames]{xcolor}
\usepackage[pagewise]{lineno}
\usepackage{natbib}
\usepackage{orcidlink}
\usepackage{comment}
\usepackage{enumitem}
\setlist[enumerate,1]{start=0}

\yil{}
\vol{}
\fpage{}
\lpage{}
\doi{}

\title{Hate Speech Detection in Turkish and Arabic: \\ A Comprehensive Study
}

\author[DEHGHAN et al.]
{
\textbf {Somaiyeh DEHGHAN$^{1,2}$\thanks{so.dehghan87@gmail.com} \orcidlink{0000-0002-5011-5821}~, 
Gökçe ULUDOĞAN$^{3}$\orcidlink{0000-0002-8684-2457}, Mehmet Umut ŞEN$^{1,2}$\orcidlink{0000-0002-8810-0502},} 
\and
\textbf{ Elif EROL$^{4}$\orcidlink{0009-0000-1554-0622},
Arzucan ÖZGÜR$^{3}$\orcidlink{0000-0001-8376-1056},
Berrin YANIKOGLU$^{1,2}$\orcidlink{0000-0001-7403-7592}} \\ [0.8em] 
$^{1}$Department of Computer Engineering, Sabanci University, Istanbul, Turkey 34956 \\ 
$^{2}$Center of Excellence in Data Analytics (VERIM), Sabanci University, Istanbul, Turkey 34956 \\ 
$^{3}$Department of Computer Engineering, Bogazici University, Istanbul, Turkey 34342 \\
$^{4}$ Hrant Dink Foundation, Istanbul, Turkey 34373 \\ [1.8em] 

\rec{} 

}

\def\E{\ifmmode{\mathbb E}\else{$\mathbb E$}\fi} 
\def\N{\ifmmode{\mathbb N}\else{$\mathbb N$}\fi} 
\def\R{\ifmmode{\mathbb R}\else{$\mathbb R$}\fi} 
\def\Q{\ifmmode{\mathbb Q}\else{$\mathbb Q$}\fi} 
\def\C{\ifmmode{\mathbb C}\else{$\mathbb C$}\fi} 
\def\H{\ifmmode{\mathbb H}\else{$\mathbb H$}\fi} 
\def\Z{\ifmmode{\mathbb Z}\else{$\mathbb Z$}\fi} 
\def\P{\ifmmode{\mathbb P}\else{$\mathbb P$}\fi} 
\def\T{\ifmmode{\mathbb T}\else{$\mathbb T$}\fi} 
\def\SS{\ifmmode{\mathbb S}\else{$\mathbb S$}\fi} 
\def\DD{\ifmmode{\mathbb D}\else{$\mathbb D$}\fi} 

\newcommand{\bse}{\begin{subequations}}
\newcommand{\ese}{\end{subequations}}
\newcommand{\ben}{\begin{enumerate}}
\newcommand{\een}{\end{enumerate}}
\newcommand{\bens}{\begin{enumerate*}}
\newcommand{\eens}{\end{enumerate*}}
\newcommand{\be}{\begin{equation}}
\newcommand{\ee}{\end{equation}}
\newcommand{\bea}{\begin{eqnarray}}
\newcommand{\eea}{\end{eqnarray}}
\newcommand{\baa}{\begin{eqnarray*}}
\newcommand{\eaa}{\end{eqnarray*}}
\newcommand{\bc}{\begin{center}}
\newcommand{\ec}{\end{center}}

\theoremstyle{corollary}

\theoremstyle{lemma}

\theoremstyle{proposition}

\theoremstyle{axiom}

\theoremstyle{conjecture}

\theoremstyle{example}

\theoremstyle{definition}

\theoremstyle{remark}

\begin{document}

\maketitle

\begin{abstract}
Online hate speech has been linked to a global rise in violence against minorities, including incidents such as mass shootings, lynchings, and ethnic cleansing. Societies grappling with this issue, particularly when hate speech targets specific groups based on religion, race, ethnicity, culture, nationality, or migration status, face the challenge of balancing freedom of expression with the need for effective content moderation on widely used online platforms. In response to this challenge, we introduce a comprehensive hate speech dataset covering five distinct topics in Turkish: refugees, the Israel-Palestine conflict, anti-Greek sentiment in Turkey, ethnic or religious communities (Alevis, Armenians, Arabs, Jews, and Kurds), and LGBTI+, alongside one topic in Arabic (refugees). In addition, we develop state-of-the-art BERT-based models to address multiple dimensions of hate speech analysis, including hate category classification, hate intensity prediction, target identification, and hate speech span detection, enabling a comprehensive understanding of hateful content in online discourse.

\end{abstract}

\section*{Disclaimer:} This study contains examples of offensive language and hate speech for research purposes. These examples do not reflect the authors’ views and are included solely to support the detection and prevention of harmful content targeting vulnerable communities.

\section{Introduction}\label{sec: introduction}
\label{Int}
With the widespread use of social media, online platforms have increasingly become spaces where hate speech can spread rapidly \citep{mathew2019spread}. Such content fosters hostility and intolerance and, in some cases, contributes to real-world violence targeting religious, racial, ethnic, and gender-based groups \citep{muller2021fanning}. As a result, detecting and mitigating hate speech has become a critical challenge for both online platforms and policymakers. To address this issue, researchers have increasingly turned to Natural Language Processing (NLP) techniques, which enable the automatic identification and analysis of hateful content in large volumes of social media text \citep{schmidt2017survey,fortuna2018survey,jahan2023systematic}.

Hate speech detection is a challenging task due to both the complexity of defining hate speech and the nuances of language. In recent years, numerous studies have focused on developing automatic methods for detecting hate speech in social media. However, there is a limited amount of research on hate speech detection in Turkish and Arabic. 
Early approaches to hate speech detection focused on using lexicons of manually chosen hateful keywords \citep{Gitari2015, Vargas2021}. However, this technique tends to be limited in effectiveness, as hate speech is not always overt. More recent research in the field, particularly for English, has shifted toward n-grams, TF-IDF, and word embedding techniques (e.g. Word2Vec, GLoVe) \citep{Coltekin2020, Sahi2018, Abro2020}. 

More recent advancements leverage large language models (e.g.,  BERT, RoBERTa, ConvBERT, mBERT, and XLM-R) for hate speech detection \citep{Husunbeyi2022, Beyhan2022, Toraman2022, Kurt2023, Dehghan2024a, Alsafari2020, Salomon2022, Khezzar2023}. Additionally, Most research in this domain adopts a binary approach to hate speech classification. However, contemporary studies have recognized the limitations of this approach, prompting a shift towards multi-class classification to gain a better understanding of the nature and dimensions of hate speech \citep{Baumler2026, Santos2026}. 

Furthermore, predicting the intensity of hate serves as a valuable metric to assess the degree of hate and offensive speech \citep{Ayele2024}. Hate intensity prediction, or hate speech strength prediction, involves assessing the level (degree) or severity (strength) of hatred expressed within a given message. This task goes beyond simply classifying a message as hateful or non-hateful; it aims to quantify how intense or extreme the hateful sentiment is. By determining the degree of hatred, this approach provides a more granular understanding of offensive content, enabling researchers and practitioners to differentiate between mild and severe expressions of hate \cite{Geleta2023, Riyadi2024, Ayele2024}. 

Despite a significant increase in hate speech detection models, many of these studies fail to address the identification of specific targets of hate speech, particularly in the Turkish and Arabic languages. Identification of the target group(s) is also essential for understanding the scope and potential impact of hate speech, as messages aimed at specific groups such as refugees, women, or LGBTI+ communities pose distinct societal risks. Unfortunately, the lack of comprehensive datasets that explicitly annotate these targets has hampered progress in this area \citep{Buyukdemirci2024, Uludogan2024b}. 

Furthermore, pinpointing the precise spans of hateful content within a text, rather than just classifying the entire message, is another promising approach that has seen limited exploration. Span detection provides more granular identification of offensive content, offering better insights into the nature and severity of hate speech \citep{Hoang2023}. Unlike traditional classification that labels entire texts, span detection identifies the specific segments responsible for the expression of hate speech. This method is particularly beneficial for human moderators, who often deal with lengthy, tedious comments and prefer clear explanations over system-generated tags that lack context. Recent studies have also demonstrated that structured explanations, such as highlighting specific hateful spans, can improve moderator efficiency by reducing decision-making time \citep{calabrese-etal-2024-explainability}.

In this study, we introduce a new dataset and perform a comprehensive analysis of hate speech discourse in Turkish and Arabic. Unlike previous works that often focus on only one or two aspects, our study encompasses hate speech classification, hate intensity prediction, target group identification, and span detection using BERT-based models. By addressing these interrelated tasks together, we aim to provide a deeper understanding of hate speech discourse, its intensity, and its targeted nature, offering a holistic perspective that advances the field. 

\noindent Our main contributions are as follows:

\begin{itemize} [itemsep=0pt]
    \item We introduce a novel and extensive hate speech dataset that covers five distinct topics—refugees, the Israel-Palestine conflict, anti-Greek sentiment in Turkey, LGBTI + and religion/race/ethnicity—in Turkish, along with one topic in Arabic (refugees).
    
    \item We develop state-of-the-art BERT-based models for multiple hate speech analysis tasks, including classification, intensity prediction, target identification, and span detection.
    
    \item We leverage ChatGPT for hashtag segmentation and synthetic data generation to improve data quality and mitigate class imbalance.
    
    \item We provide the first comprehensive benchmark for multi-dimensional hate speech analysis in Turkish and Arabic.
\end{itemize}
    
The paper is organized as follows: In Section \ref{sec:our-dataset}, we introduce our hate speech dataset, along with the data annotation process and statistical analysis of the inter-annotator agreement. In Section \ref{sec:methodology}, we detail our proposed approach, covering the architecture of our transformer-based models, the baseline model, dual contrastive learning for hate speech classification, a regression model for predicting hate speech intensity, a multi-label classification model for target identification, and token classification model for span detection, hashtag segmentation and synthetic data generation using ChatGPT. Section \ref{sec:experiment} presents our experimental results, highlighting performance gains. Finally, Section \ref{sec:conclusion} summarizes our findings and outlines future research directions.


\section{Hate Speech Dataset Introduced in This Study} \label{sec:our-dataset}

We collected data from X, formerly Twitter, on five topics in Turkish—immigrants and refugees, the Israel-Palestine conflict, anti-Greek sentiment in Turkey, ethnic or religious communities (Alevis, Armenians, Arabs, Jews, and Kurds), and LGBTI+—as well as one topic in Arabic, specifically immigrants and refugees. Our dataset partially overlaps with two publicly available datasets \citep{Beyhan2022, Arin2023}, which cover four topics: immigrants and refugees, the Israel-Palestine conflict, anti-Greek sentiment, and the Istanbul Convention. Furthermore, our dataset represents the finalized version of the one used in \citep{AnnotationFocus2025}, during a big collaborative project on the topic.
The initial set of tweets collected within this project has already been shared in \citep{Arin2023, Uludogan2024a, Dehghan2024a, Dehghan2024b, seker-2025-hatecattr}. The topics covered in the dataset are as follows:

\textit{\textbf{Immigrants and Refugees in Turkey:}}
In recent years, the civil wars in Syria and Afghanistan have driven countless individuals to seek refuge in Turkey. As of 2023, around 3.4 million Syrians\footnote{https://multeciler.org.tr/eng/number-of-syrians-in-turkey/} and approximately 300,000 Afghans\footnote{https://www.voanews.com/a/afghan-refugees-in-turkey-hope-for-relocation-fear-deportation/7400549.html}
 have settled in the country. Initially, public sentiment toward refugees was largely supportive; however, as the number of asylum seekers grew, challenges associated with their integration and disinformation about refugees receiving privileges unavailable to Turkish citizens have contributed to increasing negative attitudes. This trend is not unique to Turkey; similar shifts in public opinion have been observed globally, often resulting in heightened hostility and hate speech targeting refugees on social media platforms. 

\textit{\textbf{Israel-Palestine Conflict:}} 
The Israel–Palestine conflict, ongoing since the mid-20th century, remains one of the most complex and divisive disputes in modern history. Public discourse on the issue has long been shaped by deeply polarized opinions between pro-Israeli and pro-Palestinian groups. In Turkey, the conflict has sparked significant debate and strong reactions, underscoring its broader regional and global implications. It is important to emphasize that our dataset was collected prior to October 7, 2023, and therefore does not capture 
the subsequent escalation of hate speech.

\textit{\textbf{Anti-Greek Sentiment in Turkey:}} 
Anti-Hellenism, or anti-Greek sentiment, refers to hostility toward Greeks, Greek culture, or Greece. Relations between Turkey and Greece have long been shaped by disputes over issues such as Aegean sovereignty, territorial waters, airspace, and minority rights. In 2022, Greece's increased military presence on certain Aegean islands further escalated tensions between the two countries\footnote{https://www.dailysabah.com/politics/eu-affairs/greece-scales-up-crete-naval-base-armament-drive}. Such developments often contribute to the spread of anti-Greek rhetoric and hate speech in public discourse.

\textit{\textbf{Ethnic or religious communities (Alevis, Armenians, Arabs, Jews, and Kurds):}} 
In addition to the main topics, hate speech frequently targets ethnic and religious communities in Turkey and the Middle East. Groups such as Alevis, Armenians, Arabs, Jews, Kurds, and Roma are often subjected to discriminatory and hateful discourse shaped by historical, political, and social tensions. To capture these forms of hate speech, we included tweets targeting these communities in our dataset.

\textit{\textbf{LGBTI+:}}In Turkey and many Muslim-majority countries, negative attitudes toward LGBTI+ individuals are often influenced by cultural, religious, and social norms. Conservative interpretations of Islam and traditional views on family structures and gender roles can contribute to societal discrimination and hostility toward LGBTI+ communities. In some contexts, political rhetoric further reinforces these attitudes, increasing the prevalence of anti-LGBTI+ hate speech.

\subsection{Data Annotation} \label{sec:annotation}

We use a prescriptive annotation method to assign tweets to specific hate speech categories and a descriptive approach to rate the perceived intensity of hate speech on a scale of $[0, 10]$. The annotation guidelines were refined iteratively to reduce ambiguities in categorization. Hate speech intensity was evaluated separately from the categories, providing a secondary metric to examine its relationship with the hate speech classifications.

The tweets were grouped into batches of 50 and assigned to three annotators via Label Studio\footnote{https://labelstud.io/}. Each annotator was instructed to label the tweets individually, following the provided guidelines. To maintain label quality, each tweet was independently annotated by three different individuals, with each batch being reviewed by three separate annotators across distinct ports. Multiple selections were allowed for the “Target Group" and “Hate Speech Category" labels, as multiple groups can be targeted inside one tweet. In such cases, we split the annotator vote among the selected groups or categories. 
Our annotation guidelines are made publicly available\footnote{\label{guideurl_1}\href{https://hrantdink.org/attachments/article/4413/UTILIZING\%20AI\%20AGAINST\%20HATE\%20SPEECH\%20A\%20guide\%20to\%20annotation,\%20classification,\%20and\%20detection.pdf}{English Annotation Guideline Document, HDV Publications}}\textsuperscript{,}\footnote{\label{guideurl_2}\href{https://hrantdink.org/attachments/article/4412/NEFRET\%20S\%C3\%96YLEM\%C4\%B0YLE\%20M\%C3\%9CCADELEDE\%20YAPAY\%20ZEK\%C3\%82\%20Etiketleme,\%20s\%C4\%B1n\%C4\%B1fland\%C4\%B1rma\%20ve\%20tespit\%20k\%C4\%B1lavuzu.pdf}{Turkish Annotation Guideline Document, HDV Publications}}.
%
%
\noindent The following six categories were decided for a comprehensive labelling:

\begin{enumerate} [itemsep=0pt]
   \item \textbf{No Hate Speech:} Tweet does not contain hate speech. 
   
   \item \textbf{Exclusionary/Discriminatory Discourse:} These are narratives in which 
   an identity is portrayed as inherently inferior or undeserving compared to the dominant group regarding societal benefits, rights, and freedoms. For example, statements like “Mülteciler sağlık hizmetlerinden yararlanmasın" (Refugees should not access healthcare) and “Kadınlar iş dünyasında daha az yer almalı" (Women should have a lesser role in the workforce) exemplify discriminatory discourse.
   
   \item \textbf{Exaggeration, Generalization, Attribution, Distortion:} 
  These are discourses that generalize individual events or situations to an entire group, manipulate facts by distorting them, or attribute isolated incidents to an entire identity based on the actions of a few. For instance, expressions like
  “Yunan yine kışkırtma peşinde (The Greeks seek provocation again)" and  “Mültecilerin aldıklar\textsc{i} yardımlar yüzünden halkımız fakir kald\textsc{i} (Our people is poor because of the aid refugees receive)" following specific and unrelated incidents illustrate such manipulative and generalized hate speech.

  \item \textbf{Symbolization:} These are discourses in which an aspect of identity is employed as a means of insult, hatred, or humiliation, often symbolizing the identity in a derogatory manner. For instance, expressions like “Bizi Eurovision’da Yahudi mi temsil edecek?" (Will the Jew represent us at Eurovision?) 
  exemplify such hateful and symbolic discourse.
   
  \item \textbf{Swearing, Insult, Defamation, Dehumanization:} 
  These are discourses that involve direct profanity, insults, or contempt towards an identity, often dehumanizing them by describing their actions or characteristics with terms typically used for non-human beings. For example, “Küstah Rum’a Gözdağı" (Intimidation to the Arrogant Greek) and  “Danimarkal\textsc{i} itler iftar sofraların\textsc{i} basıp Kuran-\textsc{i} Kerim yaktılar (Danish dogs raided iftar and burned the Quran)" reflect such hateful and dehumanizing expressions.

  \item \textbf{Threat of Enmity, War, Attack, Murder, or Harm:} These are discourses that include hostile expressions towards an identity, often evoking the threat of war or violence, or expressing a desire to harm the targeted group. For example,
  “Yunanistan ateşle oynuyor; bir gece ansızın gelebiliriz! (Greece is playing with fire; we may come unexpectedly one night!)" reflects such a threatening discourse.
\end{enumerate}

\vspace{6pt}

In the following, we provide statistics for the binary and multi-class classification problems in Tables \ref{tab:turkish-data-2class}, \ref{tab:turkish-data-4class}, \ref{tab:turkish-data-6class}, and \ref{tab:arabic-data-all}. 
For multi-class categorization, we used two different settings. One of the settings involves 6 classes (fine-grained), which are the “no hate speech” class and the five categories in the Table \ref{tab:turkish-data-6class}. In the other setting, we designed a 4-class classification (coarse-grained) model by merging the categories 2 and 3 as a single category, and 4 and 5 as a single category for simplifying the task based on \cite{AnnotationFocus2025}, shown in Table \ref{tab:turkish-data-4class}. 
For training hate speech classification models, we used only instances with agreements or those where majority voting is clear for hate speech category annotation. 


\subsection{Target Annotation}
Target annotations were also assigned to each tweet in the dataset as multi-label annotations, as shown in Table \ref{tab:target}, which presents target statistics for the Turkish and Arabic datasets. The target categories are defined as follows:
\vspace{0pt}
\begin{enumerate} [itemsep=0pt]
    \item \textbf{Target group not specified or not present:} Cases where the target identity is vague or not explicitly stated.
    \item \textbf{Country, Nationality, Race, Ethnicity:}  
    Instances where an identity is targeted based on their country, nationality, race or ethnicity. This includes categories such as refugees, Greeks, Armenians, Kurds, Roma, and Arabs.
    \item \textbf{Religion:} 
    Instances where an identity face discrimination due to their religious beliefs. The relevant categories in this study include Jews and Alevis.
    \item \textbf{Gender, Sexual Orientation:} Discourses where an identity is targeted based on their gender or sexual orientation. The categories in this project include LGBTI+s and Women.
\end{enumerate}




\begin{table}[th]
\centering
\scriptsize
\caption{Statistics for 2-class labelled in Turkish tweets}
\label{tab:turkish-data-2class}
\setlength{\tabcolsep}{4pt}
\renewcommand{\arraystretch}{1.3}
\vspace{2pt}
\begin{tabular}{|l|c|c|c|c|c|c|c|c|c|c|}
\hline
  \multicolumn{1}{|c|}{} & Refugees 
& \multicolumn{1}{l|}{Isr.-Pal.} 
& \multicolumn{1}{l|}{Tr.-Gr.}  
& \multicolumn{1}{l|}{Alevis} 
 & \multicolumn{1}{l|}{Arabs} 
& \multicolumn{1}{l|}{Armenians}  
& \multicolumn{1}{l|}{Kurds} 
& \multicolumn{1}{l|}{LGBTI+} 
& \multicolumn{1}{l|}{Total} \\ \hline
0: No hate speech 
& 489    & 1700   & 938    & 685  & 409  & 201 & 733 & 523 & 5,678\\  
1: Hate speech
& 1,654  & 972    & 930    &106   & 491  & 672  & 127  & 305 & 5,257\\  \hline 
Total 
& 2,143  & 2,672  & 1,868  & 791  & 900  & 873  & 860  & 828 & 10,935\\  \hline
\end{tabular}
\end{table}
\vspace{-38pt}
\begin{table}[]
\centering
\scriptsize
\caption{Statistics for 4-class labelled in Turkish tweets.}
\label{tab:turkish-data-4class}
\renewcommand{\arraystretch}{1.3}
\setlength{\tabcolsep}{4pt}
\vspace{2pt}
\begin{tabular}{|l|c|c|c|c|c|c|c|c|c|c|}
\hline
\multicolumn{1}{|c|}{}  & Refugees & \multicolumn{1}{l|}{Isr.-Pal.} & \multicolumn{1}{l|}{Tr.-Gr.}   
& \multicolumn{1}{l|}{Alevis} 
& \multicolumn{1}{l|}{Arabs}
& \multicolumn{1}{l|}{Armenians}
& \multicolumn{1}{l|}{Kurds}
& \multicolumn{1}{l|}{LGBTI+} & \multicolumn{1}{l|}{Total} \\ \hline
0: No hate speech    
& 528   & 1,750  & 1,035   & 722   & 452  & 225 & 774 & 560 & 6,046 
\\ [0.2cm] 

\begin{tabular}[c]{@{}l@{}}1: Exclusionary/Discriminatory Discourse \end{tabular}
& 977  & 1  & 5   & 14  & 26  & 42 & 10 & 3  & 1,078  \\ [0.2cm]  

\begin{tabular}[c]{@{}l@{}}2: Exaggeration, Generalization, Attribution, \\ \hspace{7pt} Distortion, Symbolization\end{tabular}
& 198   & 185  & 278  & 19   & 215  & 341 & 42  & 74  & 1,352  \\ [0.3cm]  

\begin{tabular}[c]{@{}l@{}}3: Swearing, Insult, Defamation, \\ \hspace{7pt} Dehumanization, Threat of Enmity,
\\ \hspace{7pt} War, Attack,
Murder or Harm \end{tabular}
& 193   & 582   & 470   & 28   & 81  & 117 & 29 & 139 & 1,639  \\ [0.0cm] \hline
Total       
& 1,896  & 2,518   & 1,788  & 783  & 774 & 725 & 855 & 776 & 10,115 \\ [0cm] \hline
\end{tabular}
\vspace{0pt}
\end{table}
\vspace{-38pt}
\begin{table}[]
\centering
\scriptsize
\caption{Statistics for 6-class labelled in Turkish tweets.}
\label{tab:turkish-data-6class}
\renewcommand{\arraystretch}{1.2}
\setlength{\tabcolsep}{4pt}
\vspace{2pt}
\begin{tabular}{|l|c|c|c|c|c|c|c|c|c|c|}
\hline
\multicolumn{1}{|c|}{}  
& Refugees & \multicolumn{1}{l|}{Isr.-Pal.} & \multicolumn{1}{l|}{Tr.-Gr.}   
& \multicolumn{1}{l|}{Alevis} 
& \multicolumn{1}{l|}{Arabs} 
& \multicolumn{1}{l|}{Armenians} 
& \multicolumn{1}{l|}{Kurds} 
& \multicolumn{1}{l|}{LGBTI+} & \multicolumn{1}{l|}{Total} \\ \hline
0: No hate speech    
& 531    & 1,760   & 1,045  & 727   & 453 & 283 & 779 & 565  & 6,098  \\ [0.2cm] 
\begin{tabular}[c]{@{}l@{}}1: Exclusionary/Discriminatory Discourse \end{tabular}
& 1,064    & 1    & 6    & 16   & 39  & 63 & 15 & 3 & 1,207\\ [0.2cm] 

\begin{tabular}[c]{@{}l@{}}2: Exaggeration,
 Generalization, Attribution,
\\ \hspace{7pt} Distortion \end{tabular}
& 143    & 151    & 73    & 9   & 96 & 95 & 20 & 29 & 616  \\ [0.3cm] 

3: Symbolization  
& 5    & 16     & 182    & 5  & 64 & 167 & 5 & 21 & 465  \\ [0.2cm] 
\begin{tabular}[c]{@{}l@{}}4: Swearing, Insult, Defamation, Dehumanization\end{tabular}
& 106     & 211    & 132    & 22   & 79 & 99 & 25 & 134 & 808   \\ [0.3cm] 

\begin{tabular}[c]{@{}l@{}}5: Threat of Enmity,  War, Attack,  
 Murder or Harm \end{tabular}
& 63      & 304     & 294    & 4   & 3 & 14 & 3 & 7 & 692  \\ [0.0cm] \hline
Total       
& 1,912   & 2,443   & 1,732  & 783  & 734 & 676 & 847 & 759 & 9,886   \\  \hline
\end{tabular}
\vspace{0pt}
\end{table}
\vspace{-38pt}
\begin{table}[]
\centering
\scriptsize
\caption{Statistics for 2-class, 4-class, and 6-class labeled in Arabic tweets on \textit{Refugees} topic}
\label{tab:arabic-data-all}
\renewcommand{\arraystretch}{1.2}
\setlength{\tabcolsep}{3.5pt}
\begin{tabular}{|l|c|l|c|l|c|}
\hline
\multicolumn{1}{|c|}{2-class} & Size & \multicolumn{1}{c|}{4-class}   & Size & \multicolumn{1}{c|}{6-class} & Size \\ \hline
0: No Hate Speech   & 2,176    
& 0: No Hate Speech & 2,240    
& 0: No Hate Speech & 2,309   \\ [0.1cm]

1: Hate Speech      & 334    
& \begin{tabular}[c]{@{}l@{}}1: Exclusionary/Discriminatory \\ \hspace{9pt}Discourse \\ [0.1cm]  \end{tabular} & 4    
&\begin{tabular}[c]{@{}l@{}}1: Exclusionary/Discriminatory \\ \hspace{9pt}Discourse\end{tabular} & 8 \\ [0.1cm]

&      
& \begin{tabular}[c]{@{}l@{}}2: Exaggeration, Generalization, \\ 
\hspace{7pt} Attribution, Distortion,  Symbolization \\ [0.1cm] \end{tabular} 
& 116    
& \begin{tabular}[c]{@{}l@{}}2: Exaggeration, Generalization, \\ 
\hspace{7pt} Attribution, Distortion\end{tabular} & 69 \\ [0.1cm]

&      
& \begin{tabular}[c]{@{}l@{}}3: Swearing, Insult, Defamation, \\ 
\hspace{7pt} Dehumanization, Threat of Enmity, \\ 
\hspace{7pt} War, Attack, Murder or Harm\end{tabular} & 72    
& 3: Symbolization   & 7   \\ [0.0cm]

&      
&      
&      
& \begin{tabular}[c]{@{}l@{}}4: Swearing, Insult, Defamation, Dehumanization\end{tabular}  & 41    \\ [0.2cm]

&      
&      
&      
& \begin{tabular}[c]{@{}l@{}}5: Threat of Enmity, War,  Attack, Murder  or Harm\end{tabular}  & 10    \\ [0.0cm]\hline

Total   & 2,510    &     & 2,432    &   & 2,444    \\  \hline
\end{tabular}
\vspace{0pt}
\end{table}

\vspace{-38pt}
\begin{table}[]
\centering
\scriptsize
\caption{Statistics on hate directed towards different general targets in our Turkish and Arabic dataset}
\label{tab:target}
\renewcommand{\arraystretch}{1.2}
\setlength{\tabcolsep}{4pt}
\begin{tabular}{|l|c|c|}
\hline
Targets  & Turkish Dataset & Arabic Dataset \\  \hline
0: Target group not specified or not present
& 4,327           & 1,979         \\ 
1: Country, Nationality, Race, Ethnicity
& 6,742           & 444         \\ 
2: Religion 
& 1,351           & 227         \\ 
3: Gender, Sexual Orientation
& 216            & 119         \\   \hline
\end{tabular}
\vspace{0pt}
\end{table}
\begin{table}[] 
\centering 
\scriptsize
\caption{Distribution of tweets and spans across hate speech categories.}  \label{tab:tweet_per_category}
\renewcommand{\arraystretch}{1.2}
\setlength{\tabcolsep}{4pt}
\begin{tabular}{@{}|l|c|c|@{}} 
\hline Category & \#Tweets & \#Spans \\ \hline 
1: Exclusionary/Discriminatory Discourse & 683 & 856 \\ 
2: Exaggeration, Generalization, Attribution, Distortion & 868 & 1104 \\
3: Symbolization & 700 & 800 \\ 
4: Swearing, Insult, Defamation, Dehumanization & 1015 & 1373 \\
5: Threat of Enmity, War, Attack, Murder, or Harm & 1015 & 1396 \\ \hline 
\end{tabular}
\end{table}

\newpage
\subsection{Span Annotation}

Beyond tweet-level annotations, we also conducted span-level annotations to identify the specific segments responsible for hate speech, following the annotation methodology described in \citep{seker-2025-hatecattr}. Initial span annotations were generated using GPT-4o (gpt-4o-2024-08-06) and subsequently refined by human annotators to correct errors and remove hallucinated spans. After the manual refinement process, the final span-level dataset consists of 2,981 tweets and 4,465 hateful spans. Detailed statistics regarding category is presented in Tables~\ref{tab:tweet_per_category}.


\section{Methodology} \label{sec:methodology}
In this section, we first introduce our baseline model (Section \ref{sec:baseline}), then we introduce our dual contrastive learning model for hate speech classification (Section \ref{sec:DCL}). Then, we describe our target identification and span detection models (Sections \ref{sec:target-model} and \ref{sec:span-model}).
After that, we describe our zero-shot strategy and few-shot strategy for hashtag segmentation and synthetic data generation using ChatGPT, respectively (Sections \ref{sec:hashtag-segment}).


\subsection{Baseline Model for Hate Speech Classification Problem} \label{sec:baseline}

We design our baseline model using transfer learning, with a single layer on top of the learned encoder (BERT) \citep{Devlin2019} to predict the hate speech categories using cross-entropy loss as follows:

\begin{equation}
    L_{CE} = -\sum_{i=1}^{N} y_ilog(\hat y_i)
\end{equation}
where $y_i$ is the target value for the $i$th input and $\hat y_i$ is the prediction.

\subsection{Dual Contrastive BERT Model for Hate Speech Classification} \label{sec:DCL}

We also design our dual contrastive model using transfer learning, with a single layer on top of the learned encoder (BERT) to predict the hate speech categories as in\citep{Dehghan2024a}. 
As we use a contrastive design, the input of our model is a batch  of tweets including both hate and non-hate text (binary-class or multi-class). 
%
The model is trained using a dual contrastive loss function that is a combination of cross-entropy loss ($L_{CE}$) and supervised contrastive loss ($L_{SCL}$):

\begin{equation}
    L_{DualCL} = L_{CE} + \lambda L_{SCL}
\end{equation}

where $\lambda\in\left [ 0, 1 \right ]$ is a weighting hyperparameter that controls the impact of these two loss functions. 
We tried different $\lambda$ values and observed that the best performance for 
$\lambda = 0.5$.  
The two objectives learn classifier parameters and improve the the quality of the representations of the features, simultaneously.

As our supervised contrastive loss ($L_{SCL})$, we use the loss introduced in \citep{Khosla2021}. They extended the self-supervised NT-Xent loss \citep{Chen2020} to the fully supervised setting by adding multiple positives \textcolor{black}{from the same class}. 
The NT-Xent loss only accepts one positive sample $x_i^+$, \textcolor{black}{obtained via augmentation}, for an anchor $x_i$, and uses the other samples in the mini-batch
as negatives, obtaining a mini-batch of $(x_i,x_i^+, x_{1}^-, ..., x_{K}^-).$
Supervised contrastive loss extends the number of positives in the mini-batch to $P$ positive samples \textcolor{black}{from the training set},
obtaining a mini-batch of  $(x_i,x_{1}^+,…,x_{P}^+, x_{1}^-, ..., x_{K}^-)$. 
Our SCL is  defined as:
\begin{equation}
L_{SCL}=-\frac{1}{P}\sum_{p=1}^P\log{\frac{{e^{(x_i \cdot x_{p}^+/\tau{})}}}{\sum_{k=1}^Ke^{(x_i \cdot x_k^-/\tau{})}}}
\end{equation}

where \textcolor{black}{$x_i$ is the  embedding vector of sentence $s_i$}, $x_{p}^+$ and $x_{k}^-$ indicate positive and negative  samples respectively; the symbol $\cdot$ denotes the inner product and $\tau \in R^+$ denotes the temperature parameter set to $0.1$.

\subsection{Hate Intensity Prediction Problem} \label{sec:regression-model}

We design our regression model using transfer learning, with a single layer on top of the learned encoder (BERT) to predict the hate speech degree (intensity) using MSE loss as follows:

\begin{equation}
        L_{MSE} = \frac{1}{N}\sum_{i=1}^{N} (y_i- \hat y_i)^2
\end{equation}

where $y_i$ and $\hat y_i$ are the desired values and predicted values, respectively.

\subsection{Multi-label Target Classification Problem} \label{sec:target-model}

We fine-tune the pretrained BERT model by incorporating an additional output layer specifically for the labeling task. This layer features a feedforward neural network with sigmoid activation, enabling the model to independently assign a score (0/1) to each label. We use BCEWithLogitsLoss from PyTorch, which combines a Sigmoid layer and binary cross-entropy loss (BCE Loss). BCE loss learns the probability distribution $P(y_i|\overrightarrow{s})$, for a sentence feature vector $\overrightarrow{s}$ and the predicted label of the $i_{th}$ class, $i \in {0, 1}$, $y_i$, as follows:

\begin{equation}
    L_{BCE} = -\frac{1}{N}\sum_{i=1}^{N}\left(y_i\cdot log(p(s_i))+(1-y_i)\cdot log (1-p(s_i))\right)
\end{equation}

where the probability $p(s_i)$ is the predicted probability that
instance (sentence) $s_i$ belongs to the default class and $y_i$ is the true label, either 0 or 1, for instance $i$. 


\subsection{Span Detection Model} \label{sec:span-model}

We model span detection as a token classification problem by fine-tuning a pretrained BERT model. Specifically, we formulate two token-level tasks: (i) binary span detection and (ii) multi-class span categorization.

For binary detection, we adopt the BIO tagging scheme, where each token is labeled as beginning (B), inside (I), or outside (O) of a hateful span. For categorization, we use an IO tagging scheme, where each token is either labeled as outside (O) any hateful span or assigned to one of five hate speech categories.

We fine-tune the BERT encoder with a token-level classification layer and optimize the model using token-level Cross-Entropy (CE) loss. Given the input token sequence $x$, the model predicts token label probabilities $p(y_i|x)$ for each token $i$. The loss function is defined as:

\begin{equation} L_{TokenLevelCE} = -\frac{1}{N}\sum_{i=1}^{N}\sum_{c=1}^{C}y_{i,c}\cdot \log (p_{i,c}) \end{equation}

where $N$ is the number of tokens, $C$ is the number of classes, $y_{i,c}$ is the one-hot encoded ground truth for token $i$ and class $c$, and $p_{i,c}$ is the predicted probability of token $i$ belonging to class $c$.

\subsection{Hashtag Segmentation and Synthetic Data Generatıon Using ChatGPT} \label{sec:hashtag-segment}

Hashtags can be a rich source of information for sentiment analysis and hate speech detection, as they often encapsulate users' opinions and sentiments succinctly. People frequently use hashtags to convey their views on various topics, making them crucial for understanding the overall tone and context of social media posts. 

Since, BERT Tokenizer is not able to segment hashtags correctly, we tried other BERT based tool, namely Hashformers\footnote{https://github.com/ruanchaves/hashformers} \citep{Rodrigues2021}, which is developed specifically for hashtag segmentation. However, it also failed to correctly segment hashtags in Turkish, while it is very slow and requires a lot of memory to run (about 22 GB of GPU RAM).

Recently, ChatGPT has excelled in various NLP tasks, including text classification, sentiment analysis, and text generation. Leveraging this capability, we employed ChatGPT in a zero-shot setting for hashtag segmentation.
Since some hashtags in Turkish are written in Arabic or English (e.g. \#genocideingazza, \#babykiller), we included a step in our prompt for ChatGPT to first translate these hashtags into Turkish before segmenting them.
From our dataset covering five topics, we extracted about 3,447 unique hashtags. Tables \ref{tab:hashtag-segment-example} show an example of hashtag segmentation by these three mentioned methods.

Furthermore, recent studies have shown that Large Language Models (LLMs) can generate synthetic data that helps address low-resource challenges in NLP \citep{Li-synthetic2023, Guo2024}. Therefore, we used ChatGPT (gpt-3.5-turbo-1106) with few-shot prompting to generate additional samples for underrepresented hateful classes, including swearing, insult, threat, injury, enmity, and harm.

For each topic in our dataset (Refugees, Israel-Palestine conflict, Turkey-Greece relations, religion/ethnicity, and LGBTI+), we provided representative hateful examples and generated additional instances. We also created samples targeting \textit{women} and developed a hate lexicon based on frequently observed hate expressions. Using these resources, ChatGPT generated 761 synthetic samples, including 547 hateful instances. 

\begin{table}
\centering
\scriptsize
\caption{Hashtag segmentation example using BERTurk Tokenizer, Hashformer tool \citep{Rodrigues2021} and ChatGPT.}
\renewcommand{\arraystretch}{1.2}
\label{tab:hashtag-segment-example}
\begin{tabular}{|l|l|}
\hline
Method & \#ülkemdemülteciistemiyorum (\#Idontwantrefugeesinmycountry)                       \\  \hline
BERTurk Tokenizer    & \begin{tabular}[c]{@{}l@{}}{[}`\#', `ülke', `\#\#m', `\#\#dem', `\#\#ülte',  `\#\#ci', `\#\#istem', `\#\#iyorum'{]}\end{tabular}
\\ [0.05cm] \hline
Hashformer \citep{Rodrigues2021} & {[}`ül kemd emültec iis t e miy orum'{]}  \\  \hline
ChatGPT  & {[}`ülkem', `de', `mülteci', `istemiyorum'{]}  \\ [0.05cm] \hline
\end{tabular}
\end{table}

\section{Experimental Results} \label{sec:experiment}

In this section, we present comprehensive experiments and their results to evaluate the effectiveness of the proposed approaches in identifying and detecting hate speech across various tasks, including binary classification, multi-class classification, target identification, and span detection, and we discuss each result in detail.


\subsection{Training Setting}
We use a pretrained BERT transformer with a linear classification head on top of the pooled output, using the Huggingface Transformer package. For Turkish and Arabic, we start fine-tuning from pretrained checkpoint of BERTurk\footnote{https://huggingface.co/dbmdz/bert-base-turkish-uncased} and AraBERT\footnote{https://huggingface.co/aubmindlab/bert-base-arabert}, respectively. We allocated approximately 80\% of the data for training and 20\% for testing. 


\subsection{Experiment 1: Binary and Multi-class Classification in Turkish and Arabic: }

In our experiments, we evaluated two models, BERTurk-CE and BERTurk-DualCL, across three classification settings: binary (2-class), coarse-grained (4-class), and fine-grained (6-class) hate speech categorization. The 2-class setting focuses on detecting the presence or absence of hate speech, while the 6-class model distinguishes between ``no hate speech" and five specific hate speech categories. To create a 4-class setting, we followed the category mapping strategy proposed in \citep{AnnotationFocus2025}, where categories 2 and 3 were merged, as well as categories 4 and 5, to produce a more generalizable classification task. For all experiments, We only included samples where annotators reached agreement or where the majority vote was clear, thereby ensuring label quality. As a result of this filtering and the mapping between classification settings, the number of training and test samples varied slightly across the three experiments.

\vspace{6pt}
\noindent\textbf{\textit{Results and Discussion:}}

Table~\ref{tab:result-classification} presents the performance of the baseline Cross-Entropy (CE) and the proposed DualCL models across binary (2-class), coarse-grained (4-class), and fine-grained (6-class) hate speech classification tasks. Across all settings and both languages, DualCL consistently outperforms the CE baseline, demonstrating the benefits of combining contrastive learning with cross-entropy loss.

For the Turkish dataset, DualCL achieves Macro-F1 scores of 84.85\%, 75.35\%, and 66.66\% in the 2-, 4-, and 6-class settings, respectively, with improvements of up to 1.88 points over CE. The gains are particularly evident in the more challenging fine-grained classification task. For the Arabic dataset, the improvements are more substantial, especially in the 4-class and 6-class settings, where Macro-F1 increases from 42.31\% to 47.41\% and from 22.86\% to 31.38\%, respectively.

As expected, both models achieve their best performance in the binary setting, while performance decreases as the number of classes increases. Nevertheless, DualCL maintains a consistent advantage across all scenarios. Despite the smaller size and narrower topical focus of the Arabic dataset, the larger relative improvements suggest that DualCL is particularly effective in low-resource and domain-specific hate speech classification tasks.


\begin{table}[]
\centering
\scriptsize
\caption{Results of baseline (CE) and our DualCL models on Turkish and Arabic datasets}
\label{tab:result-classification}
\renewcommand{\arraystretch}{1.3}
\setlength{\tabcolsep}{4pt}
\begin{tabular}{|lcccccc|}
\hline
\multicolumn{1}{|c|}{\multirow{2}{*}{Model}}    & \multicolumn{2}{c|}{2-class}             & \multicolumn{2}{c|}{4-class}             & \multicolumn{2}{c|}{6-class} \\ \cline{2-7} 
\multicolumn{1}{|c|}{}                          & Macro-F1 & \multicolumn{1}{c|}{Accuracy} & Macro-F1 & \multicolumn{1}{c|}{Accuracy} & Macro-F1      & Accuracy     \\ \hline
                                                & \multicolumn{6}{c|}{Turkish Dataset}                                                                               \\ \hline
\multicolumn{1}{|l|}{BERTurk-CE (baseline)}     & 83.66    & \multicolumn{1}{c|}{83.69}     & 74.25    & \multicolumn{1}{c|}{80.35}     & 64.78         & 79.60        \\ \hline
\multicolumn{1}{|l|}{BERTurk-DualCL (proposed)} & \textbf{84.85} & \multicolumn{1}{c|}{\textbf{84.87}} & \textbf{75.35} & \multicolumn{1}{c|}{\textbf{80.55}} & \textbf{66.66} & \textbf{80.46} \\ \hline
                                                & \multicolumn{6}{c|}{Arabic Dataset}                                                                                \\ \hline
\multicolumn{1}{|l|}{AraBERT-CE (baseline)}     & 80.58    & \multicolumn{1}{c|}{92.38}     & 42.31    & \multicolumn{1}{c|}{93.25}     & 22.86         & 94.52        \\ \hline
\multicolumn{1}{|l|}{AraBERT-DualCL (proposed)} & \textbf{82.60} & \multicolumn{1}{c|}{\textbf{92.58}} & \textbf{47.41} & \multicolumn{1}{c|}{\textbf{93.86}} & \textbf{31.38} & \textbf{94.72} \\ \hline
\end{tabular}
\end{table}
\begin{table}[]
\scriptsize
\centering
\caption{Results of hate intensity (degree of hate) prediction on Turkish and Arabic dataset}
\label{tab:result-regression}
\setlength{\tabcolsep}{4pt}
\renewcommand{\arraystretch}{1.3}
\begin{tabular}{|c|c|c|}
\hline
\rule{-2.5pt}{10pt}Model   & RMSE & Pearson’s Correlation Coefficient (\textit{r}) \\ [0.1cm] \hline
\rule{-2.5pt}{10pt}BERTurk regression model (on Turkish dataset) & 1.64    & 73.41              \\ [0.1cm] \hline
\rule{-2.5pt}{10pt}AraBERT regression model (on Arabic dataset) & 1.20    & 67.38            \\ [0.1cm]\hline
\end{tabular}
\end{table}
\begin{table}[]
\scriptsize
\centering
\caption{Results of multi-label target classification on Turkish and Arabic dataset}
\label{tab:result-target}
\renewcommand{\arraystretch}{1.3}
\setlength{\tabcolsep}{4pt}
\begin{tabular}{|l|cccc|cccc|}
\hline
\multicolumn{1}{|c|}{\multirow{2}{*}{Hate Targets}}                  & \multicolumn{4}{c|}{\rule{-2.5pt}{10pt}Turkish Dataset}    & \multicolumn{4}{c|}{Arabic Dataset}   \\ [0cm] \cline{2-9} 
\multicolumn{1}{|c|}{}  & \rule{-2.5pt}{10pt}Precision & Recall & F1-score & Support & Precision & Recall & F1-score & Support \\ [0cm] \hline
\begin{tabular}[c]{@{}l@{}}\rule{-2.5pt}{10pt} 0: Target group is not  specified or not present\end{tabular} 
& 0.63     & 0.80   & 0.70     & 870
& 0.86     & 0.93   & 0.89    & 412       \\ [0.1cm]

\begin{tabular}[c]{@{}l@{}}1: Country, Nationality, Race, Ethnicity\end{tabular} 
& 0.80    & 0.84    & 0.82    & 1349 
& 0.53    & 0.82    & 0.65    & 80       \\ [0.1cm]

2: Religion
& 0.41    & 0.52     & 0.46    & 256
& 0.32    & 0.22     & 0.26    & 36      \\ [0.1cm]

\begin{tabular}[c]{@{}l@{}}3: Gender, Sexual Orientation\end{tabular}
& 0.40    & 0.47     & 0.43    & 49
& 0.61    & 0.56     & 0.58    & 25      \\ [0.1cm]

&           &        &          & 
&    &        &          &         \\
\hspace{11pt}Micro avg.
& 0.68    & 0.79     & 0.73   & 2524
& 0.76    & 0.85     & 0.81    & 553      \\  [0.1cm]

\hspace{11pt}Macro avg. 
& 0.56    & 0.66     & 0.60    & 2524
& 0.58    & 0.63     & 0.60    & 553      \\  [0.1cm]
\hspace{11pt}Weighted avg.  
& 0.69    & 0.79     & 0.74    & 2524
& 0.77    & 0.85    & 0.80    & 553      \\  [0.1cm]
\hspace{11pt}Samples avg. 
& 0.72    & 0.80     & 0.74    & 2524
& 0.80    & 0.86     & 0.82    & 553      \\ [0.1cm] \hline
\end{tabular}
\end{table}

\begin{table}[]
\centering
\scriptsize
\caption{Results of span detection on Turkish dataset }
\label{tab:result-span}
\renewcommand{\arraystretch}{1.2}
\begin{tabular}{@{}|l|c|c|c|@{}}
\hline
Task & Precision & Recall & F1 \\ \hline
Binary Span Detection & 0.57 $\pm$ 0.01 & 0.62 $\pm$ 0.02 & 0.592 $\pm$ 0.01 \\
Span Categorization (Multi-class) & 0.29 $\pm$ 0.02 & 0.34 $\pm$ 0.03 & 0.310 $\pm$ 0.02 \\ \hline
\end{tabular}
\end{table}

\subsection{Experiment 2: Hate Intensity Prediction in Turkish and Arabic} 

In this second experiment, we measured the Hate Intensity (strength or degree of hate) on a scale from 0 to 10. For evaluating our Hate Intensity Prediction model (regression model), we use RMSE (Root Mean Squared Error) and Pearson's correlation coefficient metrics. The Pearson correlation coefficient, often symbolized as \textit{ r}, is a widely used metric to assess linear correlations between two variables. It produces a value ranging from –1 to 1, indicating both the magnitude and direction of the correlation. 1 indicates a perfect positive linear relationship, -1 indicates a perfect negative linear relationship, and 0 indicates that there is no linear relationship between the variables.

\vspace{6pt}
\noindent\textbf{\textit{Results and Discussion:}}

Table~\ref{tab:result-regression} presents the performance of the regression models for hate intensity prediction on Turkish and Arabic datasets. The BERTurk model achieves an RMSE of 1.64 and a Pearson correlation of 73.41, while AraBERT obtains a lower RMSE of 1.20 with a Pearson correlation of 67.38. The lower RMSE of AraBERT indicates fewer large prediction errors, whereas the higher Pearson correlation of BERTurk shows a stronger alignment between predicted and actual intensity values.

The difference in performance may be influenced by the characteristics of the \textit{test dataset}. The Arabic test set contains fewer samples (533) compared to the Turkish test set (2,524), making its evaluation more sensitive to individual predictions. Overall, BERTurk provides more stable correlation results, while AraBERT demonstrates lower prediction error in a smaller, domain-specific setting.


\subsection{Experiment 3: Multi-label Target Classification in Turkish and Arabic} 

In the third experiment, we extended our analysis by incorporating target group identification along with hate speech classification and hatefulness intensity measurement. In addition to classifying and assessing the strength of hate on a scale from 0 to 10, we identified both the general and specific target groups within hateful tweets.
Multi-label general target classification involved determining whether a tweet targeted a group and categorizing it into four broad categories, as shown in Table \ref{tab:target}, for the Turkish and Arabic datasets.

\vspace{6pt}
\noindent\textbf{\textit{Results and Discussion:}}

Table~\ref{tab:result-target} presents the results of multi-label hate target classification for the Turkish and Arabic datasets. Overall, the model achieves stronger performance on the Turkish dataset, which contains a larger and more diverse set of samples. For Turkish, the highest F1-score is obtained for the \textit{``Country, Nationality, Race, Ethnicity"} category (0.82), followed by \textit{``Target not specified or not present"} (0.70), with an overall micro-average F1-score of 0.73. Lower performance is observed for underrepresented categories such as \textit{``Religion"} (0.46) and \textit{``Gender, Sexual Orientation"} (0.43).

For Arabic, the model performs best on \textit{``Target not specified or not present"} (0.89) and \textit{``Country, Nationality, Race, Ethnicity"} (0.65), reflecting the dataset's focus on refugee-related hate speech. Performance is lower for less represented categories, including \textit{``Religion"} (0.26) and \textit{``Gender, Sexual Orientation"} (0.58). These results suggest that class distribution and topic coverage have a significant impact on target classification performance.


\subsection{Experiment 5: Span Detection in Turkish}

In this experiment, we evaluate token-level detection and categorization of hateful spans in Turkish. We formulate two tasks: (i) binary span detection using the BIO tagging scheme, and (ii) multi-class span categorization using an IO scheme based on five hate speech categories.

We fine-tune BERTurk\footnote{\url{https://huggingface.co/dbmdz/bert-base-turkish-cased}} using the Huggingface Transformers library. The model consists of a linear classification head applied to token-level representations. Training is conducted using 5-fold cross-validation, while a separate set of 300 tweets is held out for final evaluation. Evaluation is performed using the SeqEval\footnote{\url{https://github.com/chakki-works/seqeval}} library, with token-level precision, recall, and F1 scores computed by flattening predictions and gold labels. Results are reported as the mean and standard deviation across folds.

To ensure robustness and avoid overfitting to a small evaluation set, we adopt a hybrid strategy that combines cross-validation with a held-out test set. The test set comprises 300 tweets that reached inter-annotator agreement in the first annotation round, ensuring high label reliability. Separate models are trained for detection and categorization, as the tasks differ in label granularity and class distribution, requiring different learning dynamics.

\vspace{6pt}
\noindent\textbf{\textit{Results and Discussion:}} 

Results are provided in Table~\ref{tab:result-span}. The span detection model demonstrates moderately strong performance in identifying hateful spans, with an F1 score of 59\%. In contrast, categorizing spans into specific hate speech types proves significantly more difficult, with an F1 score of 31\%. 

Additionally, 
the model performs best on explicit categories such as \textit{``Threat of Enmity, War, Attack, Murder, or Harm"} (F1 = 0.44), while struggling with more implicit and context-dependent categories like \textit{``Exclusionary/Discriminatory Discourse"} (F1 = 0.08) and \textit{``Exaggeration, Generalization, Attribution, Distortion"} (F1 = 0.17). These findings suggest that while explicit threats are more easily identifiable, subtle forms of hate speech require deeper contextual understanding and remain a challenge for the current approach.

\section{Summary and Conclusions} \label{sec:conclusion}

In this work, we introduced benchmark datasets for detecting and measuring hate speech in Turkish and Arabic and conducted extensive experiments across multiple tasks. The datasets include 10,953 Turkish tweets covering topics such as Refugees, the Israel–Palestine conflict, Anti-Greek sentiment in Turkey, Religion, Race/Ethnicity, and LGBTI+ issues, as well as 2,510 Arabic tweets mainly focused on Refugees. The data was annotated to support hate speech classification, intensity prediction, target identification, group detection, and span detection.

To ensure annotation reliability, each tweet was evaluated by three annotators, resulting in an average Krippendorff’s Alpha score of 0.234, indicating the challenges and subjectivity of hate speech annotation. To address this issue, we relied on agreement-based and majority-voted samples to ensure reliable training data.

Our findings demonstrate that the DualCL model consistently outperformed the cross-entropy baseline, with larger improvements in complex multi-class settings and the low-resource Arabic dataset. The models captured hate intensity variations effectively, while target and span detection models performed well but remained challenged by underrepresented categories and implicit hate expressions.
Overall, this study provides comprehensive benchmarks, datasets, and models for multi-dimensional hate speech analysis in Turkish and Arabic, supporting future research and applications.

\renewcommand{\bibfont}{\scriptsize}
\section*{Acknowledgments}
This article was produced within the scope of the project ``Utilizing Digital Technology for Social Cohesion, Positive Messaging and Peace by Boosting Collaboration, Exchange and Solidarity" (EuropeAid/170389/DD/ACT/ Multi), implemented by \textbf{the Hrant Dink Foundation (HDF)} in partnership with \textbf{Sabancı University} and \textbf{Boğaziçi University}, and supported by  \textbf{the European Union} and \textbf{the Friedrich Naumann Foundation}. The implementing parties are solely responsible for the content of this publication, and the views expressed herein do not necessarily reflect those of the supporters. Early versions of parts of this work were published in conference proceedings \citep{Dehghan2024a, seker-2025-hatecattr}. The present article provides a more comprehensive and integrated treatment of these earlier results. We would like to thank \textbf{Nural Özel} and \textbf{Hasan Kerem Şeker} for their valuable  assistance in data collection, transfer, and management throughout the annotation process. We also thank the HDF project team and the annotators for their valuable contributions.
\renewcommand{\bibfont}{\scriptsize}
\section*{Author Contributions}
All authors contributed to the conception and design of the study. \textbf{Somaiyeh Dehghan} and \textbf{Gökçe Uludo\u{g}an} jointly developed the methodology. \textbf{Somaiyeh Dehghan} implemented the binary and multi-class classification models, hatefulness intensity prediction models, multi-label target classification models, hashtag segmentation and synthetic data generation procedures, and conducted the related experiments. \textbf{Gökçe Uludo\u{g}an} implemented the span annotation process and span detection models and conducted the related experiments. \textbf{Mehmet Umut Şen} prepared the dataset and derived gold labels. \textbf{Somaiyeh Dehghan} wrote the main parts of the manuscript, while \textbf{Gökçe Uludo\u{g}an} authored the sections related to span detection. \textbf{Elif Erol} coordinated data annotation guidelines; \textbf{Berrin Yanıkoğlu} and \textbf{Arzucan Özgür} were the principal investigators of the project and helped revise the manuscript. All authors read and approved the final manuscript.

\section*{Declaration of Generative AI and AI-assisted Technologies}
The authors used ChatGPT to proofread the manuscript. After using this tool, the authors reviewed and edited the content as needed and assume full responsibility for the content of the publication.

\section*{Declaration of Competing Interests}
The authors have no relevant financial or non-financial interests to disclose. 
\renewcommand{\bibfont}{\tiny}
\setlength{\bibsep}{0pt}



\end{document}